# BIONIC

## Bionic Humans Using EAP as Artificial Muscles Reality and Challenges

**Yoseph Bar-Cohen**
Jet Propulsion Laboratory/Caltech
4800 Oak Grove Drive, M/S 82-105, Pasadena, CA 91109, USA
e-mail: yosi@jpl.nasa.gov   Web: http://ndeaa.jpl.nasa.gov

For many years, the idea of a human with bionic muscles immediately conjures up science fiction images of a TV series superhuman character that was implanted with bionic muscles and portrayed with strength and speed far superior to any normal human. As fantastic as this idea may seem, recent developments in electroactive polymers (EAP) may one day make such bionics possible. Polymers that exhibit large displacement in response to stimulation that is other than electrical signal were known for many years. Initially, EAP received relatively little attention due to their limited actuation capability. However, in the recent years, the view of the EAP materials has changed due to the introduction of effective new materials that significantly surpassed the capability of the widely used piezoelectric polymer, PVDF. As this technology continues to evolve, novel mechanisms that are biologically inspired are expected to emerge. EAP materials can potentially provide actuation with lifelike response and more flexible configurations. While further improvements in performance and robustness are still needed, there already have been several reported successes. In recognition of the need for cooperation in this multidisciplinary field, the author initiated and organized a series of international forums that are leading to a growing number of research and development projects and to great advances in the field. In 1999, he challenged the worldwide science and engineering community of EAP experts to develop a robotic arm that is actuated by artificial muscles to win a wrestling match against a human opponent. In this paper, the field of EAP as artificial muscles will be reviewed covering the state of the art, the challenges and the vision for the progress in future years.

## 1. Introduction

As humans live longer there is a growing need for availability of organs for transplant however shortage in donations necessitates the development of artificial alternatives.  Advances in medicine have led to the availability of artificial blood, replacement joints, heart valves, and heart-lung machines that are common implanted.  In the United States, nearly one in ten individuals is using some type of an implanted medical device [Malchesky, 2001].  Muscle is a critically needed organ and its availability in an artificial form for medical use can greatly contribute to the improvement of the quality of life of many humans.  The emergence of effective electroactive polymers (EAP) that are also known as artificial muscles can potentially address this need.  These materials are human made actuators that have the closest operation similarity to biological muscles.  While these actuation materials are far from being ready for use as implants enormous progress has been made in recent years turning them into a promising technology for consideration in medical applications.  Generally, these materials respond to electrical stimulation with a significant shape or size change and this characteristic behavior has added greatly to the list of desirable properties of polymer materials [Bar-Cohen, 2004].  The large strain response of EAP materials is increasingly attracting the attention of engineers and scientists from many different disciplines who are seeking novel applications. Experts in biomimetics are particularly excited about these materials since they can be applied to mimic the movement of biological creatures [Bar-Cohen and Breazeal, 2003; and Bar-Cohen, 2004].  Using this capability, EAP actuated robotic mechanisms are enabling engineers to create devices that were previously only imaginable in science fiction.  One such commercial product has already emerged in Dec. 2002 is a form of a Fish-Robot (Eamex, Japan) [http://www.eamex.co.jp/video/fish.wmv].  It swims without batteries or a motor and it uses EAP materials that simply bend upon stimulation.  For power it uses inductive coils that are energized from the top and bottom of the fish tank. This fish represents a major milestone for the field, as it is the first reported commercial product to use electroactive polymer actuators.

For several decades, it has been known that certain types of polymers can change shape in response to electrical stimulation. Initially, these EAP materials produced a relatively small strain. Since the beginning of the 1990s, a growing number of new EAP materials are emerging with a large strain response to electrical stimulation [Bar-Cohen, 2004]. The materials that have emerged were divided by the author into two major groups, including ionic and electronic EAP. Each of the groups and individual type of EAP materials has its advantages and disadvantages that need to be taken into account when considering applications.

In recognition of the need for international cooperation among the developers, users, and potential sponsors, the author organized the first EAP Conference on March 1-2, 1999, through SPIE International [Bar-Cohen, 1999]. This conference and was the largest ever on this subject, marking an important milestone and turning the spotlight onto these emerging materials and their potential. This SPIE ElectroActive Polymer Actuators and Devices (EAPAD) Conference is now organized annually and has been steadily growing in number of presentations and attendees. Currently, there is a website that archives related information and links to homepages of EAP research and development facilities worldwide [http://eap.jpl.nasa.gov], and a semi-annual Newsletter is issued electronically [http://ndeaa.jpl.nasa.gov/nasa-nde/lommas/eap/WW-EAP-Newsletter.html]. Further, the author edited and coauthored a reference book on EAP that has been published in [Bar-Cohen, 2001] where its 2$^{nd}$ edition was published in March 2004 [Bar-Cohen, 2004]. This book [http://ndeaa.jpl.nasa.gov/nasa-nde/yosi/yosi-books.htm] provides a comprehensive documented reference, technology user's guide, and tutorial resource, with a vision for the future direction of this field. It covers the field of EAP from all its key aspects, i.e., its full infrastructure, including the available materials, analytical models, processing techniques, and characterization methods.

In 1999, in an effort to promote worldwide development towards the realization of the potential of EAP materials the author posed an armwrestling challenge. A graphic rendering of this challenge is illustrated in Figure 1. In posing this challenge, the author sought to see an EAP activated robotic arm win against human in a wrestling match to establish a baseline for the implementation of the advances in the development of these materials. While such a challenge was intended to jump-start the research activity in this field, success in wrestling against humans will enable capabilities that are currently considered impossible. It would allow applying EAP materials to improve many aspects of our life where some of the possibilities include effective implants and smart prosthetics, active clothing, realistic biologically inspired robots and the fabrication of products with unmatched capabilities and dexterity. Decades from now one, can expect to see EAP materials used to replace damaged human muscles, i.e., making "bionic human." A remarkable contribution of the EAP field would be to see one day a handicapped person jogging to the grocery store using this technology. Recent advances in understanding the behavior of EAP materials and the improvement of their efficiency led to the point that the first armwrestling competition is planned for March 7, 2005 during the EAP-in-Action Session of the EAPAD Conference where three organizations (listed by order of announcement) have already stated their readiness for this competition:

1. SRI International, Menlo Park, CA, USA (Currently seeking the necessary funds to develop the required arm in order to compete)
2. Environmental Robots Incorporated (ERI), Albuquerque, New Mexico, USA
3. Swiss Federal Laboratories for Materials Testing and Research, EMPA, Dubendorf, Switzerland.

## 2. Nature as a Biologically-Inspiring Model

Evolution over millions of years made nature introduce solutions that are highly power-efficient and imitating them offers potential improvements of our life and the tools we use. Human desire and capability to imitate nature and particularly biology has continuously evolved and with the improvement in technology more difficult challenges are being considered. Imitation of biology may not be the most effective approach to engineering mechanisms using man-made capabilities. It is inconceivable to imaging flying with a machine that has feathers and flapping wings, where obviously a machine like that will not allow us to reach the distances and carry the loads that aircraft are doing today. The introduction of the wheel has been one of the most important inventions that human made allowing to travel great distances and perform tasks that would have been otherwise impossible within the life time of a single human being. While wheel based locomotion mechanisms allow reaching great distances and speeds, wheeled vehicles are subjected to great limitations with regards to traversing complex terrain with obstacles. Obviously, legged creatures can perform numerous functions that are far beyond the capability of an automobile. Producing legged robots is increasingly becoming an objective for robotic developers and considerations of using such robots for space applications are currently underway. Making miniature devices that can fly like a dragonfly; adhere to walls like gecko; adapt the texture, patterns, and shape of the surrounding as the octopus (can reconfigure its body to pass thru very narrow tubing); process complex 3D images in real time; recycle mobility power for highly efficient operation and locomotion; self-replicate; self-grow using surrounding resources; chemically generate and store energy; and many other capabilities are some of the areas that biology offers as a model for science and engineering inspiration. While many aspects of biology are still beyond our understanding and capability, significant progress has been made and the field of biomimetics is continuing to evolve [Bar-Cohen and Breazeal, 2003].

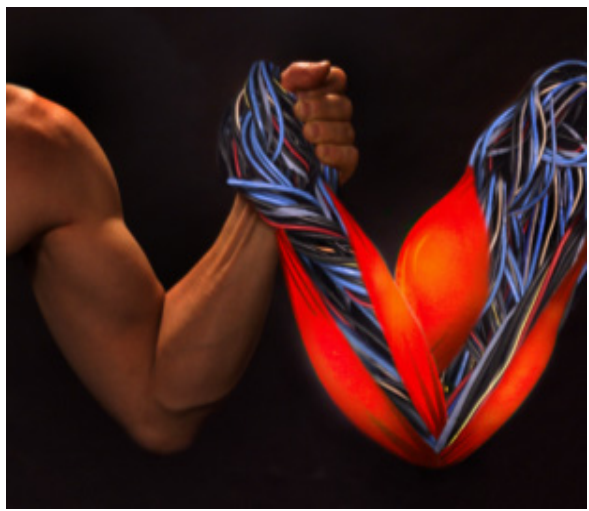

Fig. 1. Grand challenge for the development of EAP actuated robotics

The evolution in the capabilities that are inspired by biology has increased to a level where more sophisticated and demanding fields, such as space science, are considering the use of such robots. At JPL, four and six legged robots are currently being developed for consideration in future missions to such planets as Mars. Such robots include the LEMUR (Limbed Excursion Mobile Utility Robot). This type of robot would potentially perform mobility in complex terrains, perform sample acquisition and analysis, and many other functions that are attributed to legged animals including grasping and object manipulation. This evolution may potentially lead to the use of life-like robots in future NASA missions that involve landing on various to planets. The details of such future missions will be designed as a plot, commonly used in entertainment shows rather than conventional mission plans of a rover moving in a terrain and performing simple autonomous tasks. Equipped with multi-functional tools and multiple cameras, the LEMUR robots are intended to inspect and maintain installations beyond humanity's easy reach in space. This spider looking robot has 6 legs, each of which has interchangeable end-effectors to perform the required mission. The axis-symmetric layout is a lot like a starfish or octopus, and it has a panning camera system that allows omni-directional movement and manipulation operations.

## 3. EAP as Artificial Muscles

One of the key aspects of driving mechanisms that emulate biology is the development of actuators that mimic the capability of biological muscles. The potential for such actuators is continuously growing as advances are being made leading to more effective electroactive polymers (EAP) [Bar-Cohen, 2004]. These materials have functional similarities to biological muscles, including resilience, quiet operation, damage tolerance, and large actuation strains (stretching, contracting or bending). They can potentially provide more lifelike aesthetics, vibration and shock dampening, and more flexible actuator configurations. These materials can be used to make mechanical devices and robots with no traditional components like gears, and bearings, which are responsible to their high costs, weight and premature failures. Also, they could potentially be used as artificial organ to assist or operate the heart and/or its valve, the eye lid and/move the eyeball as well as control the focal length of its length, and allow mobility of the legs and/or hand as well as provide smart prosthetics (also known as cyborgs). As an example of a capability of EAP materials that is inspired by biology, the author and his JPL's NDEAA team developed a miniature robotic arm. This robotic arm illustrates some of the unique capabilities of EAP, where its gripper consists of four EAP fingers (made by Ionic polymer metal composite strips) with hooks at the bottom emulating fingernails. This arm was made to grab rocks similar to human hand.

The beginning of the field of EAP can be traced back to an 1880 experiment that was conducted by Roentgen using a rubber-band with fixed end and a mass attached to the free-end, which was charged and discharged [Roentgen, 1880]. Generally, there are many polymers that exhibit volume or shape change in response to perturbation of the balance between repulsive intermolecular forces, which act to expand the polymer network, and attractive forces that act to shrink it. Repulsive forces are usually electrostatic or hydrophobic in nature, whereas attraction is mediated by hydrogen bonding or van der Waals interactions. The competition between these counteracting forces, and hence the volume or shape change, can be controlled by subtle changes in parameters such as solvent, gel composition, temperature, pH, light, etc. The type of polymers that can be activated by non-electrical means include: chemically activated, shape memory polymers, inflatable structures, including McKibben Muscle, light activated polymers, magnetically activated polymers, and thermally activated gels [Chapter 1 in Bar-Cohen, 2004].

Polymers that are chemically stimulated were discovered over half-a-century ago when collagen filaments were demonstrated to reversibly contract or expand when dipped in acid or alkali aqueous solutions, respectively [Katchalsky, 1949]. Even though relatively little has since been done to exploit such 'chemo-mechanical' actuators, this early work pioneered the development of synthetic polymers that mimic biological muscles. The convenience and practicality of electrical stimulation and technology progress led to a growing interest in EAP materials. Following the 1969 observation of a substantial piezoelectric activity in PVF2 [http://www.ndt.net/article/yosi/yosi.htm], investigators started to examine other polymer systems, and a series of effective materials have emerged. The largest progress in EAP materials development has occurred in the last ten years where effective materials that can induce up to 380% strain have emerged [Kornbluh and Pelrine, 2004].

EAP can be divided into two major categories based on their activation mechanism including ionic and electronic. The electronic EAP are driven by Coulomb forces and they include: Dielectric EAP (shown in Fig. 2a), Electrostrictive Graft Elastomers, Electrostrictive Paper, Electro-Viscoelastic Elastomers, Ferroelectric Polymers and Liquid Crystal Elastomers (LCE). This type of EAP materials can be made to hold the induced displacement while activated under a DC voltage, allowing them to be considered for robotic applications. These materials have a greater mechanical energy density and they can be operated in air with no major constraints. However, the electronic EAP require a high activation fields (>30-V/μm) that may be close to the breakdown level. In contrast to the electronic EAP, ionic EAP are materials that involve mobility or diffusion of ions and they consist of two electrodes and an electrolyte. The activation of the ionic EAP can be made by as low as 1-2 Volts and mostly a bending displacement is induced. The ionic EAP include Carbon Nanotubes (CNT), Conductive Polymers (CP), ElectroRheological Fluids (ERF), Ionic Polymer Gels (IPG), Ionic Polymer Metallic Composite (IPMC) (shown in Fig. 2b). Their disadvantages are the need to maintain wetness and they pose difficulties to sustain constant displacement under activation of a DC voltage (except for conductive polymers).

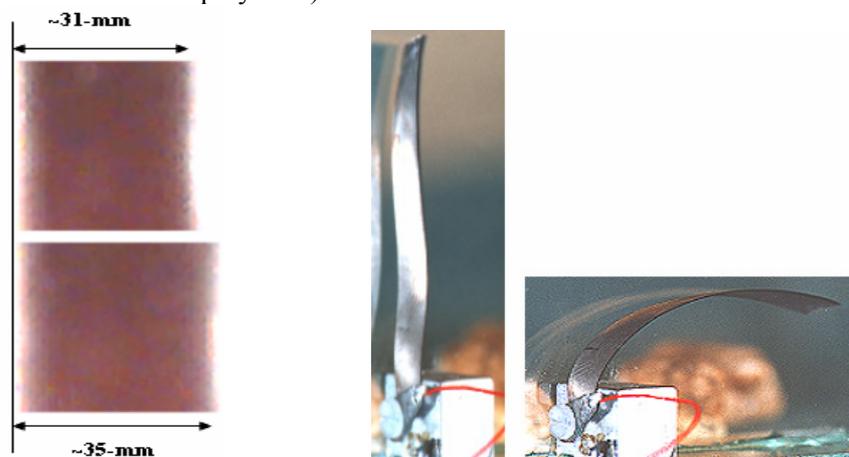

  a. Dielectric EAP in relaxed (top) and               b. IPMC in relaxed (left) and activated states (right)
  activated states (bottom)

Fig. 2. Examples of EAP materials in relaxed and activated states.

The induced displacement of both the electronic and ionic EAP materials can be designed geometrically to bend, stretch or contract. Any of the existing EAP materials can be made to bend with a significant bending response, offering an actuator with an easy to see reaction. However, bending actuators have relatively limited applications due to the low force or torque that can be induced. One important question, which has been asked by new users or researchers/engineers who are comers to this field, is the need to know where they can get these materials. This issue of unavailability of commercial EAP materials is dampening the rate of progress in the field of EAP. To help potential users, the author has established a website that describes how to make the various EAP materials [http://ndeaa.jpl.nasa.gov/nasa-nde/lommas/eap/EAP-recipe.htm]. Further, the author compiled inputs from companies that make EAP materials, prototype devices or provide

EAP related processes and services. The inputs were compiled into a handy table that is posted on one of the links of the WW-EAP webhub: http://ndeaa.jpl.nasa.gov/nasa-nde/lommas/eap/EAP-material-n-products.htm

**4. Making Robots Actuated by EAP**
With today's technology one can quite well graphically animate the appearance and behavior of biological creatures. However, in past years, engineering such biomimetic intelligent creatures as realistic robots was a significant challenge due to the physical and technological constraints and shortcomings of existing technology. Making such robots that can hop and land safely without risking damage to the mechanism, or making body and facial expression of joy and excitement are very easy tasks for human and animals to do but extremely complex to engineer. The use of artificial intelligence, effective artificial muscles and other biomimetic technologies are expected to make the possibility of realistically looking and behaving robots into more practical engineering models [Bar-Cohen and Breazeal, 2003].

To promote the development of effective EAP actuators, which could impact future robotics, toys and animatronics, two test-bed platforms were developed. These platforms are available at the Principal author's lab at JPL and they include an Android head that can make facial expressions and a robotic hand with activatable joints. At present, conventional electric motors are producing the required deformations to make relevant facial expressions of the Android. Once effective EAP materials are chosen, they will be modeled into the control system in terms of surface shape modifications and control instructions for the creation of the desired facial expressions. Further, the robotic hand is equipped with tandems and sensors for the operation of the various joints mimicking human hand. The index finger of this hand is currently being driven by conventional motors in order to establish a baseline and they would be substituted by EAP when such materials are developed as effective actuators.

The growing availability of EAP materials that exhibit high actuation displacements and forces is opening new avenues to bioengineering in terms of medical devices and assistance to humans in overcoming different forms of disability. Areas that are being considered include an angioplasty steering mechanism, and rehabilitation robotics. For the latter, exoskeleton structures are being considered to augment the mobility and functionalities of patients with weak muscles.

**5. Challenges to Developing EAP Materials as Artificial Organs**
As polymers, EAP materials can be easily formed in various shapes, their properties can be engineered and they can potentially be integrated with micro-electro-mechanical-system (MEMS) sensors to produce smart actuators. As mentioned earlier, the most attractive feature of EAP materials is their ability to emulate the operation of biological muscles with high fracture toughness, large actuation strain and inherent vibration damping. Unfortunately, the materials that have been developed so far are still exhibiting low conversion efficiency, are not robust, and there are no standard commercial materials available for consideration in practical applications. In order to be able to take these materials from the development phase to application as effective actuators, there is a need to establish adequate EAP infrastructure. Effectively addressing the requirements of the EAP infrastructure involves understanding and analytically model the ehaviour of EAP materials, as well as developing effective processing and characterization techniques.

If one considers the use of EAP as artificial organs there are challenges that need to be addressed that are common to the use of any foreign objects as implants in the human body. Such issues include biological compatibility and avoiding rejection, chemically safe use, and ability to meet the stringent functional requirements to operate as a replacement organ. Some of the issues related to the use of EAP as artificial organs include the fact that the electronic EAP group requires high voltage. At present, the materials in this group have the highest robustness and they induce the largest actuation forces however the required voltages in the range from hundreds to thousands of voltage is a concern that must be addressed. Even though the electric current is relatively low, the use of such voltage levels can cause such dangers as inducing blood clot or injury due to potential voltage breakdown. On the other hand, the ionic group of EAP materials is chemically sensitive and requires careful protection, further, it is difficult to maintain a static position because of the fact that these materials involve chemical reaction and even DC voltage causes a reaction.

Interfacing between human and machine to complement or substitute our senses may enable important capabilities for medical applications. A number of such interfaces, with some that employ EAP, were investigated or have been considered. Of notable significance is the ability to interface machines and the human brain [Wessberg et al., 2000 and Mussa-Ivaldi, 2000]. A development by scientists at Duke University enabled this possibility where electrodes were connected to the brain of a monkey, and, using brain waves, the monkey operated a robotic arm, both locally and remotely via the internet. It is envisioned that success in developing EAP actuated robotic arms that can win a wrestling match with human opponent can greatly benefit from this development by neurologists. Using such a capability to control prosthetics would require feedback to allow the human operator to "feel" the environment around the artificial limbs. Such feedback can be provided with the aid of tactile sensors, haptic devices, and other interfaces. Besides providing feedback, sensors will be needed to allow the users to monitor the prosthetics from potential damage (heat, pressure, impact, etc.) just as we are doing with biological limbs. The development of EAP materials that can provide tactile sensing is currently under way as described in [Bar-Cohen, 2004].

## 6. Summary and Outlook

Using effective EAP actuators to mimic nature would immensely expand the collection and functionality of the actuators that are currently available as well as enable making artificial organs. The prospect of developing technology that would enable making "bionic" humans with artificial muscles as science fiction TV series superhuman character is becoming increasingly feasible as the field of EAP progresses. These man-made materials operate as actuators with the closest functional similarity to biological muscles including resilience, quiet operation, damage tolerance, and large actuation strains (stretching, contracting or bending). This similarity has earned EAP the moniker Artificial Muscle and they may be used to eliminate the need for gears, bearings, and other components that complicate the construction of actuated mechanisms and are responsible to high costs, weight and premature failures. Visco-elastic EAP materials can provide more lifelike aesthetics, vibration and shock dampening, and more flexible actuator configurations.

Electroactive polymers can potentially enable bioengineering of medical applications that are considered impossible with today's technology. Important addition to this capability can be the application of telepresence combined with virtual reality using haptic interfaces. As the technology progresses, it is more realistic to expect that biomimetic prosthetics will become commonplace in our future environment. As we are inspired by biology to make more intelligent biomimetics to improve our lives we will increasingly find challenges to such implementations. The author's arm-wrestling challenge having a match between EAP-actuated robots and a human opponent highlights the potential of this technology.

## 7. Acknowledgement